\documentclass[journal]{IEEEtran}

\usepackage{xcolor,soul,framed}
\usepackage{graphicx}
\graphicspath{{.}}
\DeclareGraphicsExtensions{.pdf,.jpeg,.png,.jpg}

\usepackage[cmex10]{amsmath}
\usepackage{array}
\usepackage{mdwmath}
\usepackage{eqparbox}
\usepackage{url}
\usepackage{booktabs}
\usepackage{multirow}
\usepackage{colortbl}
\usepackage{pifont}
\let\labelindent\relax
\usepackage{enumitem}
\usepackage{hyperref}
\usepackage{newunicodechar}

\newunicodechar{̒}{} 

\definecolor{skyblue}{RGB}{70,130,180}
\definecolor{lightblue}{RGB}{135,206,250}

\begin{document}

\title{Attribute-Grounded Selective Reasoning for Artwork Emotion Understanding with Multimodal Large Language Models}

\author{Cheng~Zhang,
        Yuer~Liu,
        Zhiyu~Zhou,
        Hongxia~Xie,~\IEEEmembership{Member,~IEEE},
        and~Wen-Huang~Cheng,~\IEEEmembership{Fellow,~IEEE}%
\thanks{Cheng Zhang, Yuer Liu, Zhiyu Zhou, and Hongxia Xie are with the Department of Computer Science and Technology, Jilin University, Changchun, China. 
E-mail: Cheng Zhang (zhangcheng2122@mails.jlu.edu.cn), Yuer Liu (liuye9923@mails.jlu.edu.cn), Zhiyu Zhou (zhouzy1622@mails.jlu.edu.cn).}%
\thanks{Wen-Huang Cheng is with the Department of Computer Science, National Taiwan University, Taipei, Taiwan.}%
\thanks{Corresponding author: Hongxia Xie (e-mail: hongxiaxie@jlu.edu.cn).}}

\markboth{IEEE TRANSACTIONS ON MULTIMEDIA,~VOL.~XX, NO.~X, MARCH~2026}{Zhang \MakeLowercase{\textit{et al.}}: Attribute-Grounded Selective Reasoning for Artwork Emotion Understanding with Multimodal Large Language Models}

\maketitle

\begin{figure*}[t]
  \centering
  \includegraphics[width=1.0\textwidth]{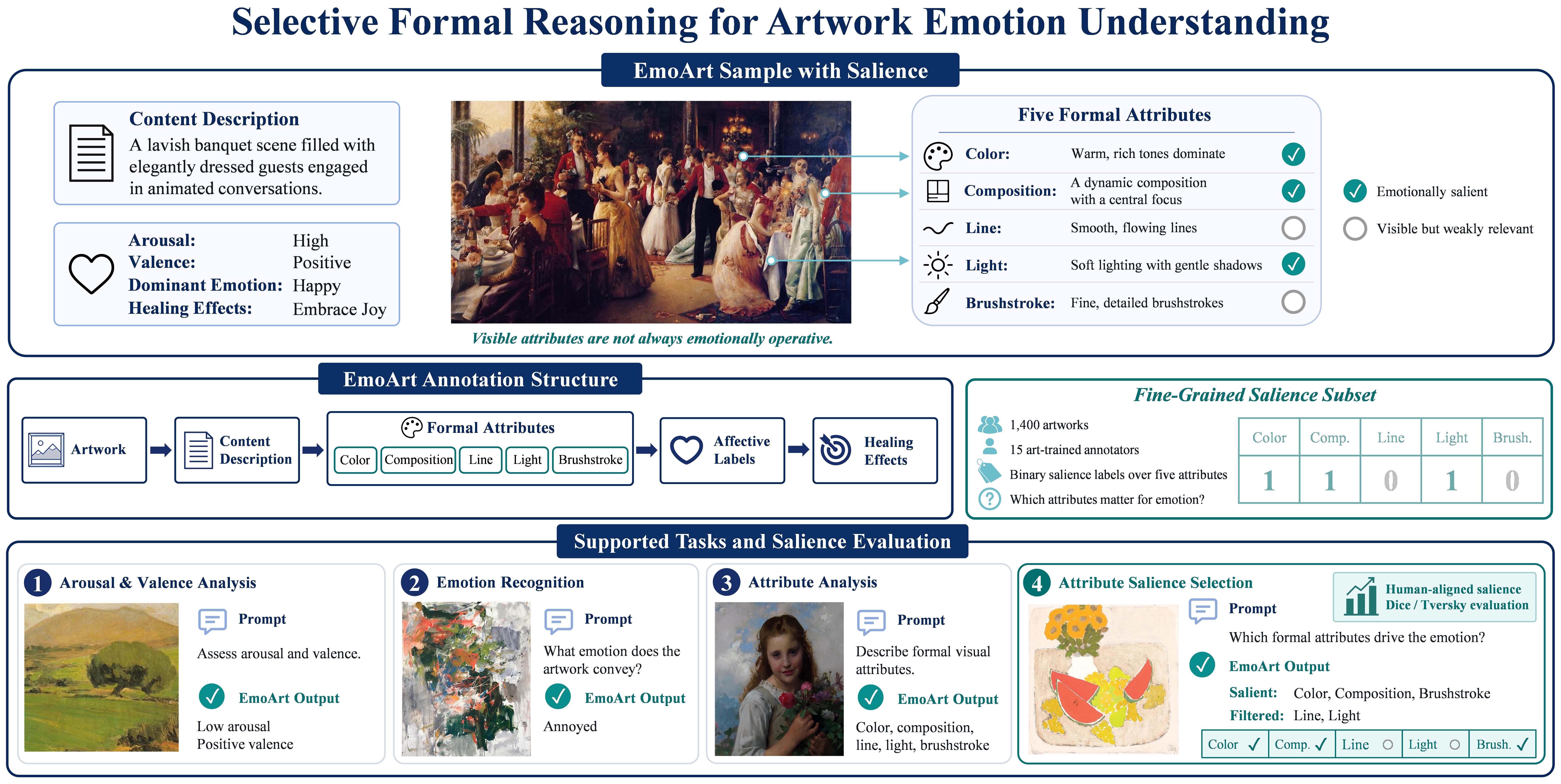}
  \caption{Overview of the EmoArt annotation structure and the proposed human salience extension. The top panel shows an EmoArt sample with content, affective labels, therapeutic effect, formal attributes, and attribute-level salience; the middle panel contrasts the original EmoArt annotation structure with the fine-grained salience subset; and the bottom panel summarizes the supported affect-prediction tasks and salience-evaluation target.}
  \label{fig:overview}
\end{figure*}

\begin{abstract}
Multimodal large language models (MLLMs) can produce fluent artwork emotion explanations, but they often suffer from \emph{attribute flooding}: they enumerate many visible formal attributes without identifying which cues actually support the affective judgment. We therefore formulate artwork emotion understanding as \emph{Attribute-Grounded Selective Reasoning} (AGSR), where predefined formal attributes serve as evidence units and only emotionally operative attributes should enter the final interpretation. To make this problem measurable, we extend \textbf{EmoArt}, originally introduced at ACM MM 2025 as a 132,664-artwork resource with content, formal-attribute, valence--arousal, and emotion annotations, by adding a 1,400-artwork human salience extension annotated by 15 art-trained annotators. This extension provides instance-level supervision for distinguishing attributes that are merely present from those that are emotionally salient. We further propose \textbf{FAB-G} (\textbf{F}ormal-\textbf{A}ttribute \textbf{B}ottleneck-\textbf{G}uided reasoning), a supervised multi-agent framework that first predicts attribute-level salience and then constrains downstream emotional analysis to the retained cues. Experiments show that FAB-G yields consistent gains in emotion, arousal, and valence prediction, achieves stronger agreement with human-marked salient attributes under Dice and Tversky metrics, and produces substantially more compact final explanations than prompting-based baselines. Cross-dataset evaluation further suggests that attribute-grounded salience selection transfers beyond the source distribution of EmoArt, while also revealing attribute-specific boundary cases. The dataset and project page are available at \href{https://zhiliangzhang.github.io/EmoArt-130k/}{\textcolor{skyblue}{https://zhiliangzhang.github.io/EmoArt-130k/}}.
\end{abstract}


\IEEEpeerreviewmaketitle

\section{Introduction}

\IEEEPARstart{A}{rtwork} emotion understanding asks a model to infer not only what an artwork depicts, but also what feeling it conveys and why. This problem is becoming increasingly important as multimodal large language models (MLLMs) are used as general interfaces for visual understanding, explanation, and open-ended reasoning~\cite{liu2023llava,openai2023gpt4v,wang2024qwen2vl}. Yet interpreting emotion in artworks remains substantially more difficult than object recognition, image captioning, or factual visual question answering. Artistic affect is rarely determined by semantic content alone: a painting may appear calm because of muted chromatic harmony, tense because of directional line structure, or melancholic because of sparse composition and subdued illumination.

This setting exposes a central limitation of current MLLMs. When asked to analyze an artwork, these models often generate fluent but overly inclusive explanations that enumerate most visible formal properties of the image. Such responses may sound plausible, but they blur the distinction between attributes that are merely \emph{present} and attributes that are genuinely \emph{emotionally salient}. We refer to this failure mode as \textbf{attribute flooding}. Under attribute flooding, a model treats many candidate cues as relevant, introducing noisy evidence and weakening interpretability. We therefore argue that artwork emotion understanding should be treated as \emph{Attribute-Grounded Selective Reasoning} (AGSR): a model should ground its explanation in a shared formal-attribute vocabulary, select the attributes that matter for a particular emotional reading, and explain the affective outcome through those selected cues.

A formalist perspective from art theory clarifies why this selectivity is necessary. Emotional and aesthetic effects are often mediated by visual form rather than by depicted subject matter alone~\cite{artbook,case2014handbook}. However, different artworks activate different subsets of formal cues. In ink painting, sparse composition and brush rhythm may dominate affect while color contributes little; in expressionist or abstract works, chromatic tension and directional structure may instead be central. A robust MLLM should therefore reason in two stages: first identifying which attributes are emotionally active, and then inferring the final emotional interpretation from those selected factors.

This distinction also matters for evaluation: a model may predict the correct emotion while still justifying it with a generic list of visible properties. AGSR captures this evidence-level quality by asking whether the cited formal cues are truly human-salient, not merely visually present.

Existing datasets only partially support this setting. General visual-emotion resources such as AffectNet~\cite{mollahosseini2017affectnet}, Emotion6~\cite{peng2015mixed}, FI~\cite{you2016building}, WEBEmo~\cite{panda2018contemplating}, and FindingEmo~\cite{mertens2024findingemo} are valuable for emotion recognition, but they primarily focus on natural images rather than on the formal analysis of artworks. Art-oriented datasets such as ArtPhoto~\cite{machajdik2010affective} and ArtEmis~\cite{achlioptas2021artemis} represent important steps toward artistic affect understanding, yet they do not explicitly annotate instance-level emotional salience over a unified set of formal attributes. As a result, they rarely indicate whether a model has selected the right artistic evidence when explaining an emotional judgment.

To make attribute flooding observable and measurable, we extend \textbf{EmoArt}~\cite{zhang2025emoart}, an ACM MM 2025 resource containing \textbf{132,664 artworks} across \textbf{56 painting styles} with structured annotations for content description, five formal visual attributes, valence--arousal, and emotion category. Specifically, we construct a \textbf{fine-grained human salience extension} of 1,400 artworks in which 15 art-trained annotators explicitly mark which attributes drive the emotional expression of each work. This extension changes the role of EmoArt from a broad affective annotation resource into an AGSR benchmark: it provides instance-level targets for both \emph{grounding} (the shared attribute vocabulary) and \emph{selection} (which attributes are emotionally operative).

We then introduce \textbf{FAB-G}, short for \textbf{F}ormal-\textbf{A}ttribute \textbf{B}ottleneck-\textbf{G}uided reasoning. FAB-G operationalizes AGSR with five attribute-specific agents that make binary salience judgments over \textit{Color}, \textit{Composition}, \textit{Line}, \textit{Light}, and \textit{Brushstroke}. Their outputs form a compact formal-attribute bottleneck, which is passed to a final analysis agent for cue-constrained emotional reasoning. This modular design directly targets attribute flooding by filtering irrelevant evidence before interpretation begins.

We evaluate this setting from multiple perspectives. Beyond final-label prediction, we measure whether a model selects the correct explanatory factors using Dice and Tversky alignment with human-marked salient attributes. We also analyze final-response length to test whether attribute-grounded selective reasoning yields more compact explanations. Finally, we perform cross-dataset transfer on re-annotated subsets from Abstract, ArtEmis, and WikiArt to examine whether the proposed framework captures reusable principles of emotional reasoning across different art-image sources rather than exploiting dataset-specific shortcuts.

The main contributions of this paper are as follows:
\begin{itemize}[leftmargin=1em, itemsep=2pt, topsep=2pt]
  \item We identify \textbf{attribute flooding} as a key failure mode of MLLM-based artwork emotion explanations and formulate \textbf{Attribute-Grounded Selective Reasoning} as an evidence-selection problem over formal attributes.
  \item We extend the ACM MM 2025 \textbf{EmoArt} resource with a \textbf{1,400-image human salience benchmark} that makes attribute flooding and AGSR measurable at the instance level.
  \item We propose \textbf{FAB-G}, a supervised formal-attribute bottleneck framework that addresses attribute flooding by decomposing artwork emotion understanding into attribute salience screening and cue-constrained emotional reasoning.
  \item We establish a multi-view evaluation protocol covering affect prediction, attribute-level reasoning alignment, final-explanation compactness, and cross-dataset transfer on Abstract, ArtEmis, and WikiArt.
\end{itemize}

\begin{figure*}[t]
  \centering
  \includegraphics[width=\textwidth]{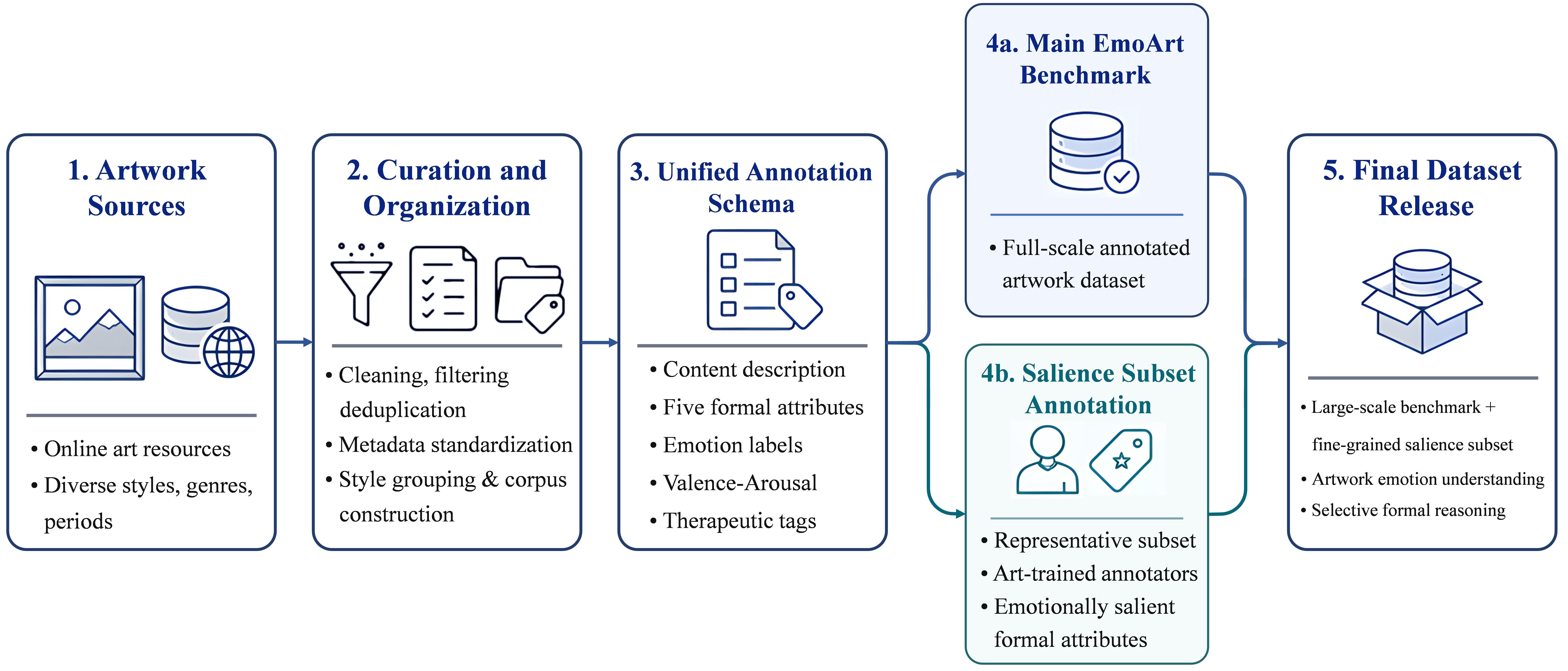}
  \caption{Pipeline for the base EmoArt resource and the supplementary salience extension.}
  \label{fig:pipeline}
\end{figure*}

\section{Related Work}

We review three lines of work most relevant to our setting: visual-emotion benchmarks, art-oriented affect understanding, and affective reasoning with multimodal large language models (MLLMs). The key distinction of this work is not the introduction of another emotion-labeling task, but the formulation of artwork emotion understanding as attribute-grounded evidence selection.

\subsection{Visual Emotion Datasets and Affective Benchmarks}

Visual-emotion datasets have evolved from controlled affective stimuli, such as \textit{IAPS}~\cite{mikels2005emotional} and \textit{GAPED}~\cite{dan2011geneva}, to larger recognition-oriented resources including \textit{Emotion6}~\cite{peng2015mixed}, \textit{FI}~\cite{you2016building}, and \textit{WEBEmo}~\cite{panda2018contemplating}. Recent benchmarks such as \textit{EmoSet}~\cite{yang2023emoset} and \textit{FindingEmo}~\cite{mertens2024findingemo} further enrich emotion taxonomies and annotation protocols. However, these resources mainly supervise \emph{what} emotion is present, while providing limited support for evaluating \emph{which visual evidence} should be used in an explanation. Building on the ACM MM 2025 EmoArt resource~\cite{zhang2025emoart}, this paper adds human-marked salience over formal attributes, shifting the benchmark from recognition alone toward explanation-oriented supervision.

\subsection{Art-Oriented Emotion Understanding}

Art-oriented datasets, including \textit{ArtPhoto}~\cite{machajdik2010affective}, \textit{ArtEmis}~\cite{achlioptas2021artemis}, and \textit{SemArt}~\cite{garcia2018semart}, show that artistic affect cannot be reduced to generic object semantics. Their labels, captions, or textual reactions provide valuable supervision for emotional plausibility and artwork understanding. Yet they do not explicitly decompose emotional interpretation into a shared set of formal evidence units such as color, composition, line, light, and brushstroke. As a result, they are less suited to testing whether a model cites the \emph{correct} artistic attributes when explaining an affective judgment. Our formulation of Attribute-Grounded Selective Reasoning (AGSR) fills this gap by treating formal attributes as supervised evidence units rather than post-hoc descriptive labels.

\subsection{MLLMs for Affective Reasoning}

Recent MLLMs such as \textit{LLaVA}~\cite{liu2023llava}, GPT-4V~\cite{openai2023gpt4v}, and \textit{Qwen2-VL}~\cite{wang2024qwen2vl} have enabled open-ended multimodal explanation, and affective models such as \textit{EmoVIT}~\cite{xie2024emovit}, \textit{AffectGPT}~\cite{lian2025affectgpt}, and \textit{AffectGPT-R1}~\cite{lian2025affectgptr1} have pushed emotion understanding toward generative reasoning. Nevertheless, fluent affective explanations can still over-activate plausible but irrelevant cues, especially in artworks where emotion is often conveyed through formal organization rather than explicit events or faces. We term this failure mode \emph{attribute flooding}. FAB-G is designed to address it by separating attribute salience screening from final emotional interpretation, enabling joint evaluation of affect prediction, evidence alignment, and final-explanation compactness.

\begin{table*}[t]
\caption{Comparison of emotion-related datasets. $R$ and $U$ denote recognition and understanding, respectively.}
\label{tab:datasets}
\renewcommand{\arraystretch}{1.2}
\centering
\begin{tabular}{lcccccccc}
\toprule
Dataset & Image Type & Label Source & Tasks & \#Image & Category & Valence\&Arousal & Attributes & Description \\
\midrule
IAPS~\cite{mikels2005emotional} & Photo & Human & R & 395 & \textcolor[HTML]{228B22}{\ding{51}} & \textcolor[HTML]{228B22}{\ding{51}} & \textcolor[HTML]{CD5C5C}{\ding{55}} & \textcolor[HTML]{CD5C5C}{\ding{55}} \\
GAPED~\cite{dan2011geneva} & Photo & Human & R & 730 & \textcolor[HTML]{228B22}{\ding{51}} & \textcolor[HTML]{228B22}{\ding{51}} & \textcolor[HTML]{CD5C5C}{\ding{55}} & \textcolor[HTML]{CD5C5C}{\ding{55}} \\
ArtPhoto~\cite{machajdik2010affective} & Art & Human & R & 806 & \textcolor[HTML]{228B22}{\ding{51}} & \textcolor[HTML]{CD5C5C}{\ding{55}} & \textcolor[HTML]{CD5C5C}{\ding{55}} & \textcolor[HTML]{CD5C5C}{\ding{55}} \\
Emotion6~\cite{peng2015mixed} & Photo & Human & R & 1980 & \textcolor[HTML]{228B22}{\ding{51}} & \textcolor[HTML]{CD5C5C}{\ding{55}} & \textcolor[HTML]{CD5C5C}{\ding{55}} & \textcolor[HTML]{CD5C5C}{\ding{55}} \\
FI~\cite{you2016building} & Photo & Human & R & 23308 & \textcolor[HTML]{228B22}{\ding{51}} & \textcolor[HTML]{CD5C5C}{\ding{55}} & \textcolor[HTML]{CD5C5C}{\ding{55}} & \textcolor[HTML]{CD5C5C}{\ding{55}} \\
WEBEmo~\cite{panda2018contemplating} & Photo & Human & R & 268K & \textcolor[HTML]{228B22}{\ding{51}} & \textcolor[HTML]{CD5C5C}{\ding{55}} & \textcolor[HTML]{CD5C5C}{\ding{55}} & \textcolor[HTML]{CD5C5C}{\ding{55}} \\
ArtEmis~\cite{achlioptas2021artemis} & Art & Human & R & 80K & \textcolor[HTML]{228B22}{\ding{51}} & \textcolor[HTML]{CD5C5C}{\ding{55}} & \textcolor[HTML]{CD5C5C}{\ding{55}} & \textcolor[HTML]{228B22}{\ding{51}} \\
EmoSet~\cite{yang2023emoset} & Photo/Art & Human\&LLM & R & 3300K & \textcolor[HTML]{228B22}{\ding{51}} & \textcolor[HTML]{CD5C5C}{\ding{55}} & \textcolor[HTML]{228B22}{\ding{51}} & \textcolor[HTML]{CD5C5C}{\ding{55}} \\
FindingEmo~\cite{mertens2024findingemo} & Photo & Human & R & 25K & \textcolor[HTML]{228B22}{\ding{51}} & \textcolor[HTML]{228B22}{\ding{51}} & \textcolor[HTML]{CD5C5C}{\ding{55}} & \textcolor[HTML]{CD5C5C}{\ding{55}} \\
\rowcolor{black!5}
EmoArt+Salience (Ours) & Art & Human\&LLM & R\&U & 132,664+1.4K & \textcolor[HTML]{228B22}{\ding{51}} & \textcolor[HTML]{228B22}{\ding{51}} & \textcolor[HTML]{228B22}{\ding{51}} & \textcolor[HTML]{228B22}{\ding{51}} \\
\bottomrule
\end{tabular}
\end{table*}

\section{Base EmoArt Resource}

\subsection{Data Collection and Filtering}

We build on EmoArt~\cite{zhang2025emoart}, originally introduced at ACM MM 2025 as a multidimensional resource for emotion-aware artistic generation. For completeness, we summarize the base resource before introducing the salience extension used in this paper. EmoArt was collected from public-domain or open-access repositories, including WikiArt, The Metropolitan Museum of Art, the National Museum of Asian Art, Japanese Print Search and Database, the National Palace Museum (Taiwan), and the National Museum of Korea, and then filtered to retain painting-centered, safe, and sufficiently represented samples. The resulting corpus contains 132,664 artworks across 56 painting styles.

\subsection{Data Annotation}
 
The base EmoArt annotations were produced with \textbf{GPT-4o}~\cite{openai2024gpt4o}, standardized prompt templates, and a multi-round verification pipeline. Each image is annotated with (1) a content description, (2) five formal attributes---brushstroke, composition, color, line, and light, (3) binary valence--arousal labels based on Russell's affect model~\cite{russell1980circumplex}, (4) one of 12 representative emotion categories, and (5) a therapeutic-effect tag inspired by art-therapy literature~\cite{case2014handbook}. The therapeutic tag is auxiliary; the AGSR benchmark and FAB-G evaluation in this paper rely on content, formal attributes, valence--arousal, emotion, and the new human salience annotations. This continuity is important because the same formal-attribute schema links the ACM MM 2025 resource, the salience extension, and the FAB-G reasoning framework.

\begin{figure}[h]
  \centering
  \includegraphics[width=\linewidth]{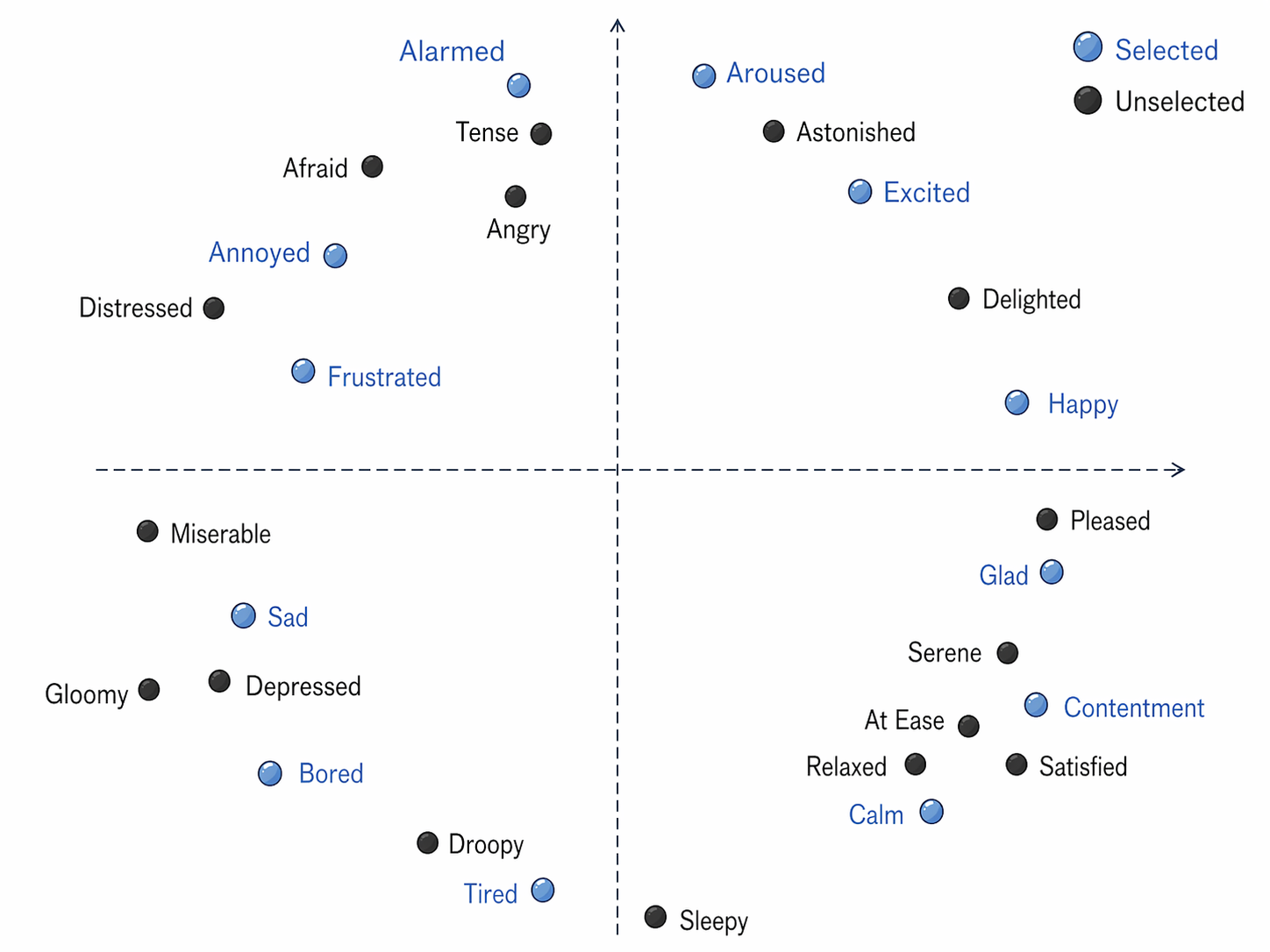}
  \caption{Distribution of 28 common emotions in the valence--arousal space and the 12 representative emotions adopted in EmoArt.}
  \label{fig:russell}
\end{figure}

This five-part schema provides the broad annotation layer, but it does not by itself indicate which attributes are emotionally operative in a particular artwork. We therefore do not treat LLM-generated annotations as a substitute for human affective judgment. Instead, they provide scalable structured metadata, while the AGSR benchmark is anchored by the human-marked salience labels introduced in Section~\ref{sec:salience_dataset}.

\begin{table}[htbp]
  \centering
  \caption{Inter-annotator agreement across annotation sections.}
  \label{tab:agreement}
  \renewcommand{\arraystretch}{1.2}
  \begin{tabular}{lcccc}
    \toprule
    \textbf{Section} & \textbf{\shortstack{True\\Prop.}} & \textbf{\shortstack{Percent\\Agree.}} & \textbf{\shortstack{Gwet's\\AC1}} & \textbf{\shortstack{Sample\\Size}} \\
    \midrule
    Description & 98.01\% & 94.25\% & 0.928 & 5,922 \\
    Visual Attributes & 98.56\% & 95.25\% & 0.944 & 5,922 \\
    Emotion & 91.47\% & 85.25\% & 0.785 & 5,922 \\
    \bottomrule
  \end{tabular}
\end{table}

\subsection{Human Validation}

Human validation was conducted on a stratified subset covering all 56 styles. Ten trained annotators assessed description quality, visual attributes, and emotion labels; Table~\ref{tab:agreement} reports 5,922 section-level validation judgments. GPT-4o~\cite{openai2024gpt4o} annotations maintain high agreement with human judgments: 98.01\% for descriptions, 98.56\% for visual attributes, and 91.47\% for emotion. These results support EmoArt as a structured base resource, while the harder question of \emph{which} attributes are emotionally operative is handled by the human salience extension.

\section{Data Analysis}

\subsection{Distribution Analysis}
\begin{figure}[h]
  \centering
  \includegraphics[width=\linewidth]{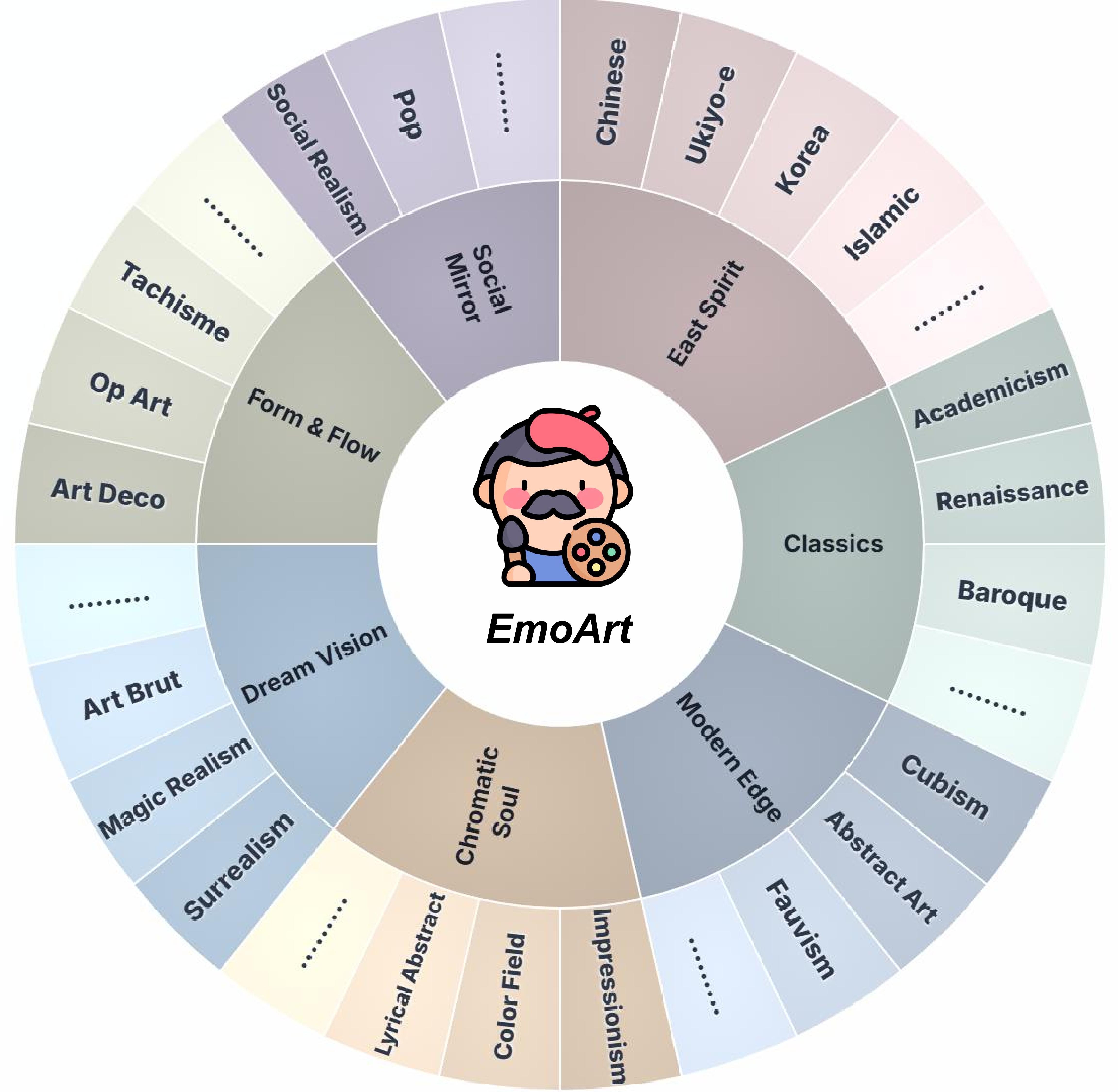}
  \caption{Distribution of the major categories and subcategories in EmoArt. The inner ring shows major categories, and the outer ring shows subcategories.}
  \label{fig:sunburst}
\end{figure}

The base EmoArt resource contains \textbf{132,664 samples} spanning \textbf{56 painting styles} across diverse historical periods and cultural contexts. For presentation clarity, we group these styles into seven thematic domains, while Figure~\ref{fig:sunburst} summarizes the dataset hierarchy. In terms of affective distribution, the dataset is dominated by \textbf{positive valence} and \textbf{low arousal}, with \textit{Calm}, \textit{Excited}, and \textit{Contentment} appearing most frequently. This tendency reflects the prevalence of soothing and uplifting emotional expression in the collected artworks, while the full dataset still preserves sufficient diversity for both generation and understanding tasks. More importantly, the breadth of styles and emotional profiles ensures that the subsequent salience benchmark is grounded in heterogeneous artistic conditions rather than in a narrow visual niche.

\section{EmoArt Salience Extension}
\label{sec:salience_dataset}

The ACM MM 2025 EmoArt resource provides broad formal-attribute annotations, but attribute flooding shows that broad attribute availability is not enough for faithful MLLM reasoning. We therefore extend EmoArt with a human salience layer built from 1,400 carefully selected artworks. The goal is to identify which formal attributes are emotionally operative in each instance, thereby turning attribute flooding from a qualitative failure mode into a measurable supervision target.

\subsection{Motivation and Problem Definition}

In the base EmoArt formulation, the five formal attributes provide a general annotation scaffold for artistic affect. However, directly supplying this fixed inventory to an MLLM can introduce redundant evidence: the model discusses all available attributes regardless of whether they are relevant to the artwork under analysis. The salience extension is designed to expose this attribute flooding behavior by asking annotators to mark only the attributes that substantively drive the emotional reading.

\subsection{Annotation Protocol}

To capture artwork-specific emotional salience, we recruit 15 annotators with formal art-training backgrounds. Instead of assigning only global emotion labels, annotators mark which formal dimensions serve as meaningful carriers of affect in a given image. The extension therefore shifts supervision from \textit{what attributes exist} to \textit{which attributes matter}, which is precisely the distinction that current MLLMs often fail to maintain.

This protocol is intentionally conservative. Annotators make selective decisions over the fixed EmoArt attribute vocabulary rather than writing open-ended rationales, so false-positive attributes and missed salient attributes can be measured directly across models and artworks.

\subsection{Dataset Scope and Annotation Targets}

The salience extension contains 1,400 representative artworks sampled with attention to style diversity, emotional diversity, and formal variation. For each artwork, the annotation target is a binary salience decision over five formal attributes: \textit{Color}, \textit{Composition}, \textit{Line}, \textit{Light}, and \textit{Brushstroke}. These labels provide direct supervision for AGSR by separating \emph{attribute grounding} from \emph{attribute selection}: the vocabulary defines what can be cited, while the salience labels define what should be cited. This annotation layer is the key component that turns EmoArt into a benchmark for validating reasoning quality rather than label prediction alone.

\section{FAB-G: Formal-Attribute Bottleneck-Guided Reasoning}

Having made attribute flooding measurable through human salience labels, we propose \textbf{FAB-G} (\textbf{F}ormal-\textbf{A}ttribute \textbf{B}ottleneck-\textbf{G}uided reasoning), a supervised multi-agent framework that operationalizes attribute-grounded selective reasoning for artwork emotion understanding. FAB-G decomposes affective reasoning into two sequential stages: \textit{attribute salience screening} and \textit{cue-constrained emotional reasoning}. The first stage determines which formal attributes are emotionally active for a given artwork, and the second stage predicts affect labels under the resulting bottleneck. This decomposition directly addresses attribute flooding by preventing the final agent from reasoning over the full inventory of visible attributes.

In implementation, all agents are instantiated with the same MLLM backbone. Each attribute agent is fine-tuned to output a binary salience decision for one formal attribute, using the artwork image and an attribute-specific instruction as input. The final analysis agent is then fine-tuned to take the image together with the predicted salience mask and to output the emotion category, arousal label, valence label, and a cue-constrained explanation. This design turns salience into an explicit supervised bottleneck rather than an implicit prompting convention.

FAB-G therefore exposes an intermediate decision variable that can be supervised, inspected, and evaluated. This is central to the method: the framework improves not only final affect labels, but also the traceability of the visual evidence that supports those labels.

\begin{figure*}[t]
  \centering
  \includegraphics[width=0.96\textwidth]{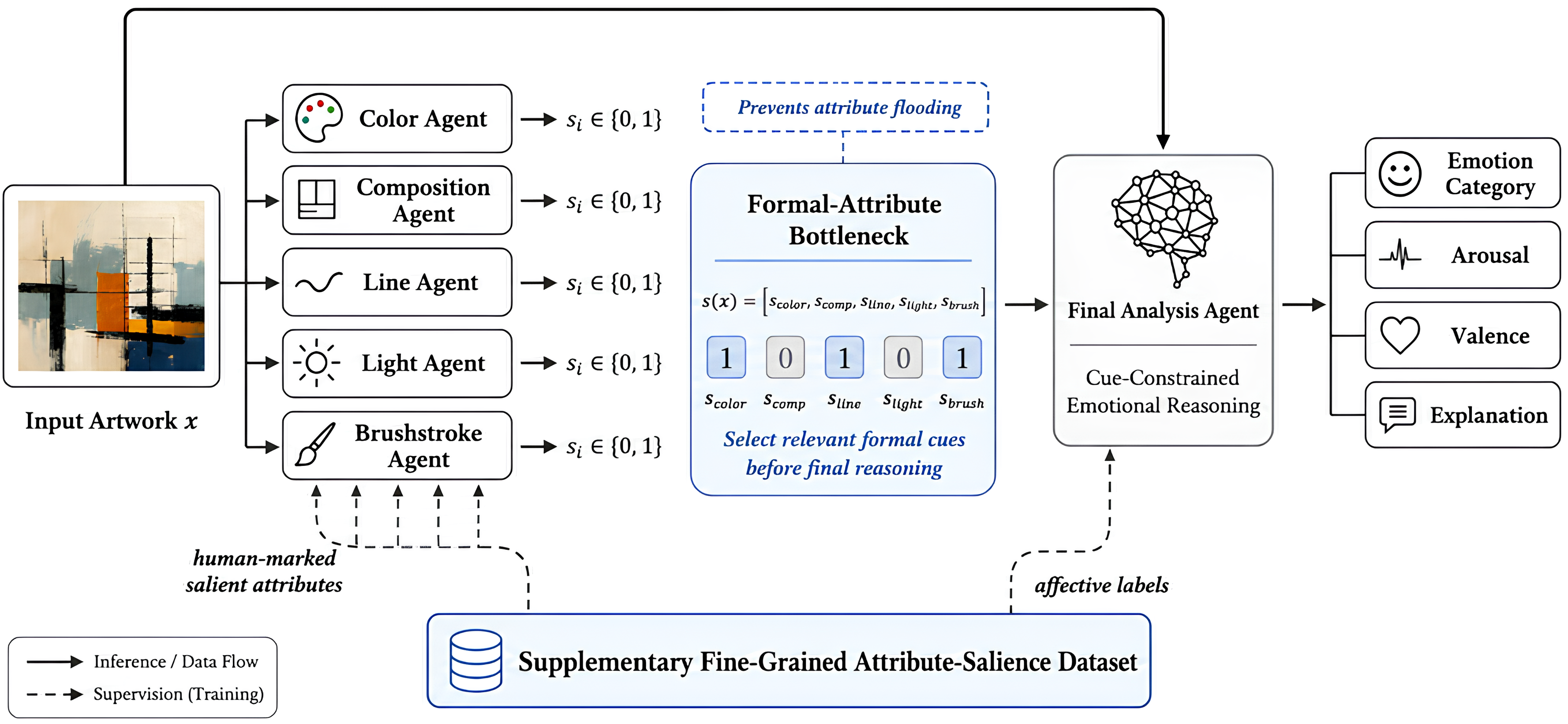}
  \caption{Overview of FAB-G. Five attribute-specific agents predict attribute salience, their outputs are aggregated into a formal-attribute bottleneck, and a final analysis agent performs cue-constrained emotional reasoning. The EmoArt salience extension supervises both the attribute agents and the final analysis agent.}
  \label{fig:fabg_arch}
\end{figure*}

\subsection{Problem Setup}

Let $x \in \mathcal{X}$ denote an input artwork, and let
\begin{equation}
\mathcal{A}=\{\mathrm{color},\mathrm{composition},\mathrm{line},\mathrm{light},\mathrm{brushstroke}\}
\end{equation}
be the predefined formal-attribute vocabulary. For each artwork, the downstream affective target is
\begin{equation}
\mathbf{y}=(y_{\mathrm{emo}}, y_{\mathrm{aro}}, y_{\mathrm{val}}),
\end{equation}
where $y_{\mathrm{emo}}$ is the emotion category, and $y_{\mathrm{aro}}$ and $y_{\mathrm{val}}$ are the binary arousal and valence labels, respectively.

The EmoArt salience extension provides instance-level supervision:
\begin{equation}
\mathcal{D}_{\mathrm{sup}}=\{(x_i, \mathbf{s}_i^{*}, \mathbf{y}_i^{*})\}_{i=1}^{N},
\end{equation}
where $\mathbf{s}_i^{*} \in \{0,1\}^{|\mathcal{A}|}$ is the human-annotated salience vector and $\mathbf{y}_i^{*}$ denotes the affective labels inherited from EmoArt. Under AGSR, $\mathcal{A}$ functions as an explicit evidence vocabulary, while $\mathbf{s}_i^{*}$ specifies which evidence units are emotionally admissible for instance $x_i$. FAB-G uses this dataset to learn both salience prediction and downstream affective reasoning.

\begin{figure*}[t]
  \centering
  \includegraphics[width=0.98\textwidth]{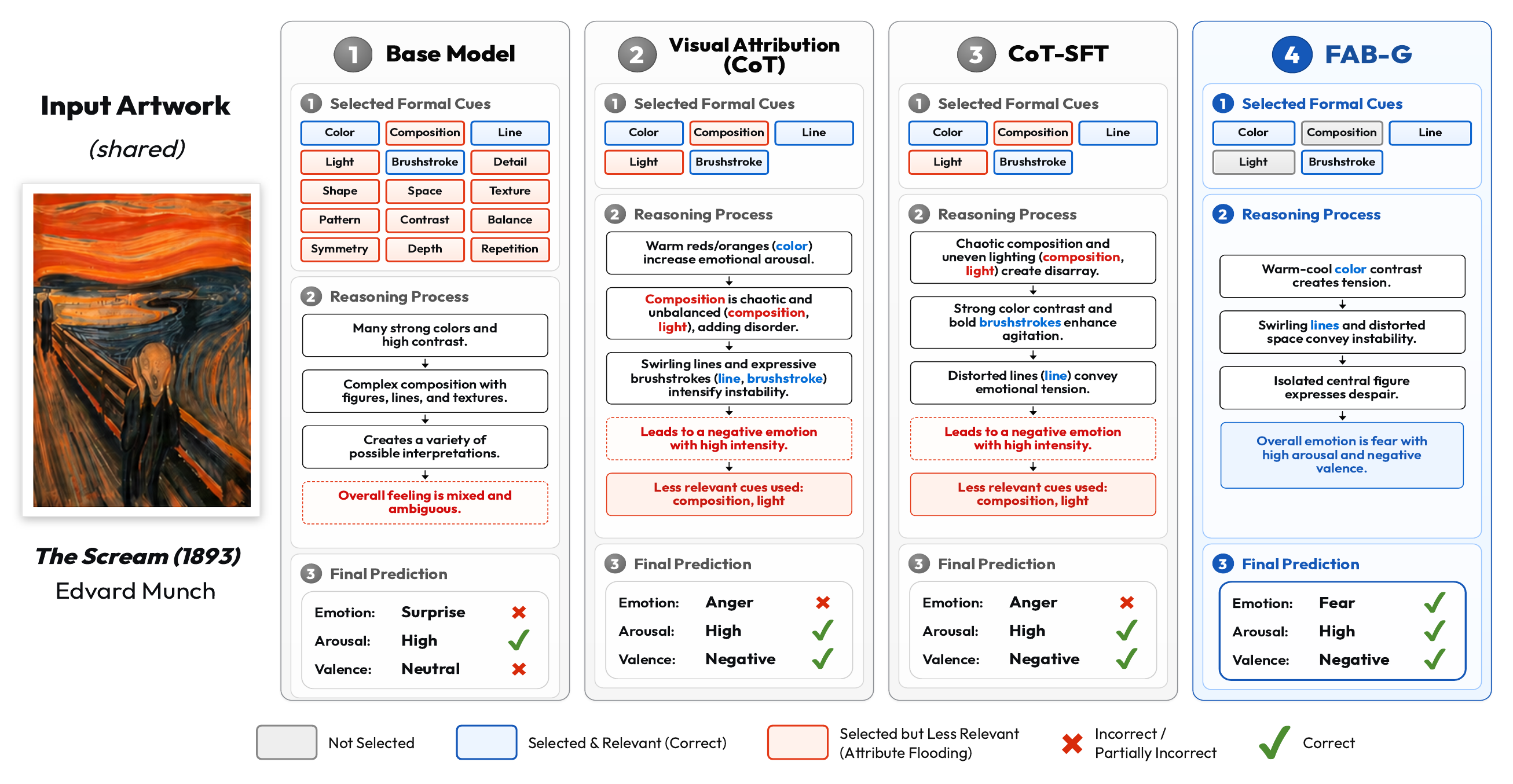}
  \caption{Qualitative case study of attribute flooding and bottleneck-guided reasoning. Baseline methods activate a broad set of visible attributes, whereas FAB-G retains a smaller set of emotionally operative cues and grounds the final prediction in those cues.}
  \label{fig:qualitative_sample}
\end{figure*}

\subsection{Attribute Salience Screening}

FAB-G instantiates one attribute agent $g_a$ for each $a \in \mathcal{A}$. Each agent solves an attribute-specific binary salience prediction problem:
\begin{equation}
s_a = g_a(x;\theta_a) \in \{0,1\},
\end{equation}
where $\theta_a$ denotes the learned parameters of agent $g_a$, and $s_a=1$ indicates that attribute $a$ is emotionally active for artwork $x$. The corresponding formal-attribute bottleneck is
\begin{equation}
\mathcal{S}(x)=\{a \in \mathcal{A} \mid s_a=1\},
\end{equation}
and its vector form is
\begin{equation}
\mathbf{s}(x)=[s_a]_{a \in \mathcal{A}} \in \{0,1\}^{|\mathcal{A}|}.
\end{equation}

Each attribute agent is trained independently using human salience labels:
\begin{equation}
\theta_a^{*}=\arg\min_{\theta_a} \sum_{i=1}^{N} \ell_{\mathrm{bin}}\big(g_a(x_i;\theta_a), s_{i,a}^{*}\big),
\end{equation}
where $\ell_{\mathrm{bin}}$ is a binary classification loss and $s_{i,a}^{*}$ is the ground-truth salience label of attribute $a$ for artwork $x_i$. The bottleneck is therefore learned directly from instance-level human supervision rather than induced by prompting alone.

The design objective of this stage is evidence selection, not forced sparsity for its own sake. In most artworks, only a subset of the five formal attributes is emotionally operative, so FAB-G is expected to operate in the regime
\begin{equation}
\|\mathbf{s}(x)\|_0 \ll |\mathcal{A}|,
\end{equation}
so that downstream affective reasoning depends on a compact subset of emotionally operative formal cues.

\subsection{Cue-Constrained Emotional Reasoning}

After salience screening, FAB-G routes the predicted bottleneck to a final analysis agent $f$. The second stage is defined as
\begin{equation}
(\hat{y}_{\mathrm{emo}}, \hat{y}_{\mathrm{aro}}, \hat{y}_{\mathrm{val}}, \hat{e}) = f(x, \mathbf{s}(x); \phi),
\end{equation}
where $\phi$ denotes the learned parameters of the final agent and $\hat{e}$ is the generated explanation. This stage is trained on the same salience extension, conditioned on the human-annotated salience vector:
\begin{equation}
\phi^{*}=\arg\min_{\phi} \sum_{i=1}^{N} \ell_{\mathrm{task}}\big(f(x_i, \mathbf{s}_i^{*};\phi), \mathbf{y}_i^{*}\big),
\end{equation}
where $\ell_{\mathrm{task}}$ denotes the downstream affective prediction loss. At inference time, the gold salience vector $\mathbf{s}_i^{*}$ is replaced by the predicted bottleneck $\mathbf{s}(x)$.

For compactness, let $\hat{\mathbf{y}}=(\hat{y}_{\mathrm{emo}},\hat{y}_{\mathrm{aro}},\hat{y}_{\mathrm{val}})$. The overall inference process can then be interpreted through the factorization
\begin{equation}
p(\hat{\mathbf{y}}, \hat{e} \mid x) \approx p\big(\mathbf{s}(x) \mid x\big)\, p\big(\hat{\mathbf{y}}, \hat{e} \mid x, \mathbf{s}(x)\big),
\end{equation}
which makes explicit the AGSR modeling assumption that affective interpretation should be conditioned on selected attribute evidence rather than on the full attribute inventory.

\subsection{Training and Inference Perspectives}

The two stages of FAB-G are trained under complementary supervision. The attribute agents learn to approximate the human-marked salience distribution, whereas the final analysis agent learns to map salience-constrained evidence to affective outcomes. This separation yields a clear division of labor: the first stage determines \emph{which attribute evidence is admissible}, and the second stage determines \emph{how that evidence supports the final affective judgment}.

At inference time, the formal-attribute bottleneck serves as an intermediate control variable that restricts the evidence space available to the final agent. Without this bottleneck, an MLLM may discuss color, composition, line, light, and brushstroke indiscriminately, even when several dimensions are incidental to the emotional reading. FAB-G instead forces the reasoning trajectory to pass through a compact set of emotionally salient factors.

\subsection{Why the Bottleneck Improves Reasoning}

FAB-G is not a training-free prompting heuristic. It is a supervised framework in which both salience screening and final affective reasoning are learned from the human salience extension of EmoArt. Let $\mathcal{S}^{\star}(x)$ denote the human-marked salient-attribute set for artwork $x$. From this perspective, attribute flooding corresponds to the regime in which
\begin{equation}
|\mathcal{S}(x) \setminus \mathcal{S}^{\star}(x)|
\end{equation}
is large, meaning that many false-positive attributes are admitted into the reasoning chain. FAB-G is designed to reduce this discrepancy before final prediction, which directly motivates the use of overlap-sensitive metrics such as Dice and Tversky in the experiments.

This design yields three practical advantages. First, final predictions are conditioned on a cleaner evidence set. Second, explanations become more auditable because each stated reason can be traced to an explicit salience decision. Third, the final response becomes more compact because the final agent no longer needs to discuss all candidate attributes exhaustively. These advantages correspond directly to our evaluation protocol: prediction accuracy measures outcome quality, Dice and Tversky scores measure evidence alignment, and final-response length measures explanation compactness.

\section{Evaluation of MLLM-Based Artwork Emotion Understanding}

With the EmoArt salience extension and the FAB-G framework, we evaluate whether the proposed formulation resolves attribute flooding in artwork emotion understanding. The evaluation focuses on three questions: whether attribute-grounded selection improves affect prediction, whether the selected evidence agrees with human-marked salient attributes, and whether the final explanation becomes more compact.
\begin{table*}[t]
  \centering
  \caption{Main results on artwork emotion understanding with multimodal large language models. Avg. Acc. denotes the unweighted mean of the three accuracy columns. Boldface indicates the best result within each model block, and the shaded row denotes FAB-G.}
  \label{tab:main_mllm_results}
  \renewcommand{\arraystretch}{1.15}
  \small
  \setlength{\tabcolsep}{6pt}
  \begin{tabular}{llcccc}
    \toprule
    \textbf{Model} & \textbf{Setting} & \textbf{Emotion Acc.}$\uparrow$ & \textbf{Arousal Acc.}$\uparrow$ & \textbf{Valence Acc.}$\uparrow$ & \textbf{Avg. Acc.}$\uparrow$ \\
    \midrule
    \multirow{5}{*}{\textbf{Qwen3-VL}} 
      & Base Model & 29.33\% & 56.67\% & 80.00\% & 55.33\% \\
      & Visual Attribution (CoT) & 29.33\% & 64.67\% & 85.33\% & 59.78\% \\
      & One-Shot + CoT & 37.33\% & 80.00\% & 85.33\% & 67.55\% \\
      & CoT-SFT & 48.67\% & 78.00\% & 88.00\% & 71.56\% \\
      \rowcolor{black!8} & \textit{Ours (FAB-G)} & \textbf{50.00\%} & \textbf{82.00\%} & \textbf{90.00\%} & \textbf{74.00\%} \\
    \midrule
    \multirow{3}{*}{\textbf{Kimi-VL}} 
      & Base Model & 26.00\% & 62.00\% & 87.33\% & 58.44\% \\
      & Visual Attribution (CoT) & \textbf{32.67\%} & 70.67\% & \textbf{89.33\%} & 64.22\% \\
      & One-Shot & 30.67\% & \textbf{80.67\%} & 85.33\% & \textbf{65.56\%} \\
    \midrule
    \multirow{3}{*}{\textbf{Qwen2.5-VL}} 
      & Base Model & 18.67\% & 68.67\% & 82.67\% & 56.67\% \\
      & Visual Attribution (CoT) & \textbf{36.00\%} & \textbf{76.67\%} & \textbf{85.33\%} & \textbf{66.00\%} \\
      & One-Shot & 26.67\% & 73.33\% & 82.00\% & 60.67\% \\
    \midrule
    \multirow{3}{*}{\textbf{MiniCPM-o-2.6}} 
      & Base Model & 29.33\% & 74.67\% & \textbf{77.33\%} & 60.44\% \\
      & Visual Attribution (CoT) & \textbf{35.33\%} & \textbf{78.00\%} & 75.33\% & \textbf{62.89\%} \\
      & One-Shot & 31.33\% & 76.00\% & \textbf{77.33\%} & 61.55\% \\
    \bottomrule
  \end{tabular}
\end{table*}

\begin{table*}[t]
  \centering
  \caption{Reasoning-quality results measured by Dice and Tversky scores. Rows are grouped by model and method. Each entry is reported as Dice / Tversky, where the Tversky index uses $\alpha=0.8$ and $\beta=0.2$. Higher is better.}
  \label{tab:reasoning_quality}
  \renewcommand{\arraystretch}{1.28}
  \footnotesize
  \setlength{\tabcolsep}{3.8pt}
  \resizebox{\textwidth}{!}{%
  \begin{tabular}{llccccccc}
    \toprule
    \textbf{Model} & \textbf{Method} & \textbf{Sample Mean}$\uparrow$ & \textbf{Attr. Mean}$\uparrow$ & \textbf{Color} & \textbf{Comp.} & \textbf{Line} & \textbf{Light} & \textbf{Brush.} \\
    \midrule
    \multirow{4}{*}{\textbf{Qwen3-VL}} & CoT & 0.7249 / 0.6954 & 0.7065 / 0.6731 & 0.6561 / 0.5973 & 0.8914 / 0.8699 & 0.6457 / 0.5409 & 0.7176 / 0.8288 & 0.6218 / 0.5286 \\
    & One-Shot & 0.6578 / 0.6204 & 0.5005 / 0.5120 & 0.8692 / 0.9083 & 0.9619 / 0.9844 & 0.0449 / 0.0293 & 0.5235 / 0.5620 & 0.1031 / 0.0762 \\
    & SFT & 0.8536 / 0.8398 & 0.8323 / 0.8136 & 0.9672 / 0.9578 & 0.9686 / 0.9586 & 0.7143 / 0.6579 & 0.7290 / 0.7588 & 0.7826 / 0.7347 \\
    \rowcolor{black!8} & \textbf{FAB-G} & \textbf{0.8742 / 0.8808} & \textbf{0.8450 / 0.8454} & \textbf{0.9492 / 0.9589} & \textbf{0.9859 / 0.9818} & \textbf{0.7683 / 0.7464} & \textbf{0.7339 / 0.7547} & \textbf{0.7879 / 0.7850} \\
    \midrule
    \multirow{2}{*}{\textbf{Qwen2.5-VL}} & CoT & 0.7091 / 0.6879 & 0.6085 / 0.6118 & 0.8465 / 0.8349 & 0.9395 / 0.9456 & 0.5124 / 0.4155 & 0.6961 / 0.8311 & 0.0482 / 0.0318 \\
    & One-Shot & 0.6755 / 0.6542 & 0.5276 / 0.5542 & 0.8353 / 0.8981 & 0.9619 / 0.9844 & 0.1895 / 0.1297 & 0.6257 / 0.7427 & 0.0253 / 0.0161 \\
    \midrule
    \multirow{2}{*}{\textbf{Kimi-VL}} & CoT & 0.7863 / 0.7700 & 0.7627 / 0.7418 & 0.8909 / 0.8909 & 0.8855 / 0.8542 & 0.6471 / 0.5670 & 0.7515 / 0.8540 & 0.6387 / 0.5429 \\
    & One-Shot & 0.5925 / 0.5249 & 0.4420 / 0.4162 & 0.6989 / 0.6298 & 0.9583 / 0.9787 & 0.1099 / 0.0729 & 0.4174 / 0.3834 & 0.0253 / 0.0161 \\
    \midrule
    \multirow{2}{*}{\textbf{MiniCPM-o-2.6}} & CoT & 0.7423 / 0.7298 & 0.6729 / 0.6708 & 0.8597 / 0.8621 & 0.9583 / 0.9787 & 0.4878 / 0.4000 & 0.7253 / 0.8684 & 0.3333 / 0.2446 \\
    & One-Shot & 0.5718 / 0.5119 & 0.4357 / 0.4194 & 0.4756 / 0.3947 & 0.9619 / 0.9844 & 0.1111 / 0.0731 & 0.5793 / 0.6122 & 0.0506 / 0.0323 \\
    \bottomrule
  \end{tabular}
  }
\end{table*}

\begin{table*}[t]
  \centering
  \caption{Cross-dataset generalization results on three external art-image sources. Overall metrics are listed before the attribute-level results. Each reasoning-quality entry is reported as Dice / Tversky. ``Acc.'' denotes emotion-prediction accuracy under the aligned label space, while ``---'' indicates that no directly comparable accuracy value is reported for that source.}
  \label{tab:cross_dataset_generalization}
  \renewcommand{\arraystretch}{1.18}
  \footnotesize
  \setlength{\tabcolsep}{4.2pt}
  \resizebox{\textwidth}{!}{%
  \begin{tabular}{lcccccccc}
    \toprule
    \textbf{Source} & \textbf{Sample Mean}$\uparrow$ & \textbf{Attr. Mean}$\uparrow$ & \textbf{Acc.}$\uparrow$ & \textbf{Color} & \textbf{Comp.} & \textbf{Brush.} & \textbf{Line} & \textbf{Light} \\
    \midrule
    \textbf{Abstract} & 0.8828 / 0.8895 & 0.6936 / 0.6876 & --- & 0.9897 / 0.9836 & 0.9048 / 0.9314 & 0.8462 / 0.8333 & 0.7273 / 0.6897 & 0.0000 / 0.0000 \\
    \textbf{ArtEmis} & 0.8912 / 0.8876 & 0.7012 / 0.6923 & \textbf{47.6\%} & 0.9812 / 0.9754 & 0.9123 / 0.9201 & 0.8215 / 0.8102 & 0.7456 / 0.7012 & 0.0012 / 0.0009 \\
    \textbf{WikiArt} & 0.8712 / 0.8654 & 0.6812 / 0.6745 & 46.9\% & 0.9723 / 0.9689 & 0.8891 / 0.9012 & 0.8012 / 0.7923 & 0.7123 / 0.6789 & 0.0005 / 0.0002 \\
    \bottomrule
  \end{tabular}%
  }
\end{table*}

\subsection{Task Definition and Evaluation Setting}

We formulate the task as \textbf{Attribute-Grounded Selective Reasoning} over artworks. Given an input painting, the model predicts the discrete \textit{emotion category}, binary \textit{arousal}, and binary \textit{valence}, while producing a natural-language explanation grounded in emotionally operative formal attributes. Success therefore depends on both affect-label correctness and evidence alignment.

Our experiments compare four representative MLLMs: \textbf{Qwen3-VL}~\cite{bai2025qwen3vl}, \textbf{Kimi-VL}~\cite{kimi2025kimivl}, \textbf{Qwen2.5-VL}~\cite{bai2025qwen25vl}, and \textbf{MiniCPM-o-2.6}~\cite{openbmb2025minicpmo26}. For each model, we evaluate a family of prompting or training strategies on the same held-out evaluation split. The \textbf{Base Model} performs emotional analysis directly from the image. \textbf{Visual Attribution (CoT)} explicitly prompts the model to analyze the artwork through major formal attributes before making a prediction. \textbf{One-Shot} or \textbf{One-Shot + CoT} provides an in-context demonstration of the desired analysis trajectory, allowing the model to imitate a structured reasoning process. For Qwen3-VL, we additionally include a \textbf{CoT-SFT} baseline, which fine-tunes the model with chain-of-thought-style supervision. Finally, \textbf{FAB-G} uses Qwen3-VL as the backbone for the five attribute agents and the final analysis agent. This comparison traces a progression from unconstrained reasoning, to attribute-aware prompting, to supervised reasoning, and finally to AGSR with explicit salience filtering.

We report \textbf{Emotion Accuracy}, \textbf{Arousal Accuracy}, and \textbf{Valence Accuracy} for affect prediction. Dice and Tversky measure agreement with human-marked salient attributes, and final-response length measures explanation compactness.

\begin{table*}[t]
  \centering
  \caption{Final-response compactness comparison. Rows are grouped by model and method. Lower is better. The remaining columns report the average final-response length conditioned on the corresponding prediction being correct.}
  \label{tab:token_efficiency}
  \renewcommand{\arraystretch}{1.18}
  \small
  \setlength{\tabcolsep}{7pt}
  \begin{tabular}{llcccc}
    \toprule
    \textbf{Model} & \textbf{Method} & \textbf{Avg. Tokens}$\downarrow$ & \textbf{Emotion Correct} & \textbf{Arousal Correct} & \textbf{Valence Correct} \\
    \midrule
    \multirow{3}{*}{\textbf{Qwen3-VL}} & CoT & 145.71 & 150.98 & 145.16 & 145.11 \\
    & One-Shot & 140.15 & 141.45 & 139.68 & 139.58 \\
    \rowcolor{black!8} & \textbf{FAB-G} & \textbf{57.56} & \textbf{61.34} & \textbf{57.85} & \textbf{58.25} \\
    \midrule
    \multirow{2}{*}{\textbf{Qwen2.5-VL}} & CoT & 106.31 & 105.20 & 107.81 & 105.14 \\
    & One-Shot & 145.48 & 144.25 & 144.03 & 143.92 \\
    \midrule
    \multirow{2}{*}{\textbf{Kimi-VL}} & CoT & 136.29 & 140.04 & 136.26 & 135.67 \\
    & One-Shot & 162.39 & 164.07 & 163.31 & 161.98 \\
    \midrule
    \multirow{2}{*}{\textbf{MiniCPM-o-2.6}} & CoT & 108.26 & 110.58 & 106.82 & 108.74 \\
    & One-Shot & 107.60 & 108.89 & 107.81 & 107.24 \\
    \bottomrule
  \end{tabular}
\end{table*}

\subsection{Main Results}

Table~\ref{tab:main_mllm_results} presents the main comparison. Attribute-aware prompting generally improves over direct image-to-emotion prediction, but prompt-level guidance remains limited. On Qwen3-VL, CoT-SFT reaches 48.67\% emotion accuracy, 78.00\% arousal accuracy, and 88.00\% valence accuracy, while FAB-G further improves to \textbf{50.00\%}, \textbf{82.00\%}, and \textbf{90.00\%}, respectively. The consistent gain over a supervised CoT baseline indicates that the benefit is not simply due to longer reasoning, but to explicitly selecting which formal cues should enter the final interpretation. The remaining gap between valence accuracy and fine-grained emotion accuracy also shows that artwork emotion understanding is not solved by stronger general-purpose MLLMs alone.

This comparison also clarifies the role of the salience extension. Prompting can tell an MLLM to consider formal attributes, but FAB-G changes the evidence set before final reasoning, making the improvement attributable to supervised attribute selection rather than to additional textual instructions.

\subsection{Qualitative Comparison}

Figure~\ref{fig:qualitative_sample} illustrates the same pattern qualitatively: baselines activate a broader set of visible attributes, whereas FAB-G retains fewer emotionally salient cues and grounds the prediction in them. This example shows how the formal-attribute bottleneck suppresses attribute flooding in practice.

\subsection{Reasoning Quality Evaluation}

Beyond final-label prediction, we evaluate whether the attributes activated by a model's explanation match the ground-truth salient attributes in the EmoArt salience extension. We use the \textbf{Dice score} and the \textbf{Tversky score} to quantify this alignment.

Let $P$ denote the set of attributes predicted in the reasoning process and $G$ denote the ground-truth salient attribute set. The Dice score is defined as
\begin{equation}
\mathrm{Dice}(P,G)=\frac{2|P \cap G|}{|P|+|G|}.
\end{equation}
Dice measures overlap between predicted and ground-truth attributes.

The Tversky score generalizes Dice by assigning asymmetric penalties to false positives and false negatives:
\begin{equation}
\mathrm{Tversky}(P,G)=\frac{|P \cap G|}{|P \cap G|+\alpha |P \setminus G|+\beta |G \setminus P|}.
\end{equation}
We set $\alpha=0.8$ and $\beta=0.2$ to penalize false positives more strongly, since attribute flooding mainly manifests as irrelevant attributes entering the reasoning chain.

We report \textbf{Sample-wise Mean}, \textbf{Attribute-wise Mean}, and per-attribute scores for \textit{Color}, \textit{Composition}, \textit{Line}, \textit{Light}, and \textit{Brushstroke}. Each cell in Table~\ref{tab:reasoning_quality} is reported as \textit{Dice / Tversky}.

Table~\ref{tab:reasoning_quality} shows that FAB-G achieves the best reasoning quality, reaching \textbf{0.8742 / 0.8808} on the sample-wise mean and \textbf{0.8450 / 0.8454} on the attribute-wise mean. Compared with Qwen3-VL (SFT), it improves both instance-level alignment and attribute-level consistency, confirming that the gain comes from selecting more human-aligned evidence rather than mentioning more attributes. The per-attribute results further show that \textit{Line}, \textit{Light}, and \textit{Brushstroke} are harder than \textit{Color} and \textit{Composition}, while FAB-G remains competitive on these subtler cues.

\subsection{Final-Response Compactness Analysis}

We also examine \textbf{final-response compactness}, measured as the length of the final generated explanation rather than total system-level inference cost. This metric captures whether the delivered explanation is less cluttered by irrelevant attributes after FAB-G performs binary attribute filtering.

As shown in Table~\ref{tab:token_efficiency}, FAB-G produces final explanations of \textbf{57.56} tokens on average, more than 50\% shorter than Qwen3-VL prompting baselines and \textbf{45.9\%} shorter than Qwen2.5-VL (CoT). The reduction also holds on correctly predicted samples, indicating that FAB-G improves explanation compactness without relying on failed or incomplete responses.

\subsection{Cross-Dataset Generalization}

To examine whether FAB-G generalizes beyond EmoArt, we evaluate on \textbf{Abstract}, \textbf{ArtEmis}, and \textbf{WikiArt}. For each source, we sample \textbf{200 artworks}, re-annotate them with the same five-attribute schema, and report Dice/Tversky reasoning quality plus emotion accuracy where compatible labels are available.

Table~\ref{tab:cross_dataset_generalization} shows strong sample-wise reasoning quality on all three external sources: \textbf{0.8828 / 0.8895} on Abstract, \textbf{0.8912 / 0.8876} on ArtEmis, and \textbf{0.8712 / 0.8654} on WikiArt. This indicates that the formal-attribute bottleneck does not merely fit the EmoArt distribution. Transfer is nevertheless attribute-dependent: \textit{Color} and \textit{Composition} are most stable, while \textit{Light} remains difficult in abstract or flattened artworks. On datasets with aligned emotion labels, FAB-G further reaches \textbf{47.6\%} on ArtEmis and \textbf{46.9\%} on WikiArt.

\subsection{Analysis of Attribute Flooding}

The results provide consistent evidence for the attribute flooding hypothesis. Prompting methods benefit from organizing reasoning around formal attributes, but FAB-G further improves over CoT-SFT by 1.33 points on emotion accuracy, 4.00 points on arousal accuracy, and 2.00 points on valence accuracy while reducing irrelevant attribute discussion. The gains indicate that human salience supervision supports a substantive modeling choice, not merely a prompt-engineering artifact.

\section{Conclusion}

This paper studies \textbf{artwork emotion understanding with multimodal large language models} through \textbf{Attribute-Grounded Selective Reasoning} (AGSR). We identify \textit{attribute flooding} as a key failure mode, where models describe many visible formal attributes without selecting those that truly support the affective judgment. To make this measurable, we extend the ACM MM 2025 EmoArt resource with a 1,400-image human salience benchmark and propose \textbf{FAB-G}, a supervised formal-attribute bottleneck framework that separates salience screening from cue-constrained emotional reasoning. Experiments show consistent gains in emotion, arousal, and valence prediction, stronger Dice/Tversky alignment with human-marked salient attributes, more compact explanations, and transferable reasoning quality across external art-image sources. These results show that explicit attribute-grounded selection improves both the outcomes and faithfulness of MLLM-based artwork emotion understanding.

The broader implication is that MLLM-based affective analysis should not be evaluated only by final-label correctness, since explanatory evidence is part of the task in artworks. By grounding reasoning in selected formal cues, AGSR offers a practical step from fluent visual description toward verifiable multimodal interpretation. Limitations remain: the current schema covers only five formal attributes, while affective interpretation may also involve iconography, cultural background, and viewer experience; the salience extension is modest relative to the full corpus; and \textit{Light} remains difficult to transfer reliably. Future work may extend AGSR with larger human-validated salience sets, culturally adaptive salience modeling, and richer schemas combining formal, semantic, and contextual evidence.
\section*{Acknowledgment}

This work was funded by the National Natural Science Foundation of China (Grant No. 62406126).

\bibliographystyle{IEEEtran}
\bibliography{IEEEabrv,Bibliography}

\end{document}